\documentclass[runningheads]{llncs}
\usepackage{graphicx}
\usepackage{comment}
\usepackage{inputenc}
\usepackage{newtxmath}
\usepackage{mathtools}
\usepackage[noend]{algpseudocode}
\usepackage{algorithm,algorithmicx}
\usepackage{amsmath}
\usepackage{graphicx,wrapfig}
\usepackage{booktabs}
\usepackage{array}
\usepackage{adjustbox}
\usepackage{listings}
\usepackage{url}
\usepackage{color}
\usepackage{xcolor}
\usepackage{enumerate}
\usepackage{lipsum}
\usepackage{pbox}
\usepackage{wasysym}
\usepackage{float}
\usepackage{paralist}
\usepackage[inline, shortlabels]{enumitem}
\usepackage{subfiles}
\usepackage{atbegshi,afterpage}
\usepackage{balance}
\usepackage{booktabs} 
\usepackage{enumitem}
\usepackage{subfig}
\usepackage{comment}
\usepackage{xspace}
\usepackage[skip=7pt]{caption}
\usepackage{hyperref}
\usepackage{cleveref}
\usepackage{pifont}
\usepackage{tabularx}
\usepackage{makecell}

\newcommand{\vadalog}{\textsc{Vadalog}\xspace}
\newcommand{\chase}{\textsc{chase}\xspace}

\newcommand*\Let[2]{\State #1 $\gets$ #2}
\algrenewcommand\algorithmicindent{0.3cm}%
\algrenewcommand\algorithmicrequire{\textbf{Precondition:}}
\algrenewcommand\algorithmicensure{\textbf{Postcondition:}}
\algnewcommand{\algorithmicgoto}{\textbf{go to}}%
\algnewcommand{\Goto}[1]{\algorithmicgoto~\ref{#1}}%
\algnewcommand\algorithmicforeach{\textbf{for each}}
\algdef{S}[FOR]{ForEach}[1]{\algorithmicforeach\ #1\ \algorithmicdo}
\algrenewcommand\alglinenumber[1]{\scriptsize #1:}
\begin{document}
\title{Fine-tuning Large Enterprise Language Models via Ontological Reasoning}
%
%
\author{Teodoro Baldazzi\inst{1} \and
Luigi Bellomarini\inst{2} \and
Stefano Ceri\inst{3} \and
Andrea Colombo\inst{3} \and\\
Andrea Gentili\inst{2} \and
Emanuel Sallinger\inst{4,5}}
\authorrunning{T. Baldazzi et al.}
%
\institute{Universit\`a Roma Tre, Italy \and
Banca d'Italia, Italy \and
Politecnico di Milano, Italy \and
TU Wien, Austria \and
University of Oxford, UK}
\maketitle              
\begin{abstract}
Large Language Models (LLMs) exploit fine-tuning as a technique to adapt to diverse goals, thanks to task-specific training data. Task specificity should go hand in hand with domain orientation, that is, the specialization of an LLM to accurately address the tasks of a given realm of interest. However, models are usually fine-tuned over publicly available data or, at most, over ground data from databases, ignoring business-level definitions and domain experience. On the other hand, Enterprise Knowledge Graphs (EKGs) are able to capture and augment such domain knowledge via ontological reasoning. With the goal of combining LLM flexibility with the domain orientation of EKGs, we propose a novel neurosymbolic architecture that leverages the power of ontological reasoning to build task- and domain-specific corpora for LLM fine-tuning.
\end{abstract}
%
%
%
%
\section{Introduction: Context and Overview of the Approach}
\label{sec:introduction}
With the recent soar of AI-based chatbots, currently led by OpenAI's ChatGPT, the field of Natural Language Processing (NLP) and, in particular, Large Language Models (LLMs), faced a major turning point and transcended its relevance in academia and industry, steering the attention of the general public towards generative AI.
While many approaches are being proposed that exploit powerful pre-trained LLMs, such as T5~\cite{RaffelSRLNMZLL20} and GPT~\cite{radford2018improving}, to address a plethora of industrial tasks, current solutions show limited effectiveness at specializing the models on enterprise domains, from finance to genomics.
In our community, such domain-specific knowledge can be captured by combining factual data from corporate databases with business-level definitions as ontologies in \textit{Enterprise Knowledge Graphs} (EKGs), and further augmented via \textit{ontological reasoning}.
In this paper, we build upon this domain representation and propose a novel solution to accurately specialize LLMs on core enterprise NLP tasks.

\smallskip \noindent
\textbf{Limits of task-specific fine-tuning.} 
LLMs can be pre-trained on extensive datasets and, often, specialized with a \textit{fine-tuning} process that customizes them so as to perform given NLP tasks~\cite{ruder2019transfer}, such as \textit{question-answering}, \textit{language translation}, \textit{named-entity recognition}, \textit{document summarization}, \textit{sentiment analysis}, and more~\cite{brown2020language}.
According to a very common usage pattern, general-purpose LLMs are fine-tuned for a specific NLP task based on extensive cross- or domain-generic textual corpora that are publicly available~\cite{rae2021scaling}.

While this approach highlights good generalization capabilities and a surprising human-style interaction, the obtained models have major shortcomings in that they lack enterprise knowledge and trivially fail to solve domain-specific NLP tasks.
For instance, in the financial domain, state-of-the-art yet generalist models have shown poor performance for different NLP tasks, for which, on the other hand, further fine-tuning with large additional text corpora has been proved to be helpful in improving the results, such as in the case of \textit{FinBert}~\cite{liu2021finbert}.

\smallskip \noindent
\textbf{Limits of domain-specific fine-tuning.}
Going further, recent studies are exploring the usage of factual data from enterprise databases to fine-tune LLMs and try to tackle domain-specific question-answering tasks: the factual information is leveraged to synthesize prompt-response pairs based on the data and customize the LLM in a task- and domain-specific direction. 
A primary example is the \textit{SKILL project}~\cite{2022-skill}, where an LLM is directly trained on factual triples derived from the translation into natural language---the so-called \textit{verbalization}---of Wikidata (namely, the \textit{KELM corpus}~\cite{agarwal2020knowledge}) for question-answering tasks. 
Similarly, other approaches highlight possible improvements of accuracy in question-answering tasks, when textual information is first captured into a database, which is then verbalized and employed for fine-tuning~\cite{Andrus2022}.

Yet, even the combination of general-purpose knowledge of the pre-trained model and the domain data still offers an accuracy that is not acceptable for core tasks in specialized domains.
For example, \textit{BloombergGPT}~\cite{DBLP:journals/corr/abs-2303-17564} is an LLM fine-tuned on a wide range of financial data, combining internal enterprise knowledge with publicly-available datasets.
The results show that the model fine-tuned for the question-answering task outperforms state-of-the-art counterparts by being able to correctly answer questions related to the financial domain.
However, \textit{BloombergGPT} has been tested only on questions whose answers are already contained in (or directly entailed by) the factual information of the input databases, either as data or meta-data (e.g., schema information). 
It is reasonable, in fact, that it does not have enough fine-tuning data or logical capabilities to go further. 

\smallskip \noindent
\textbf{A look beyond current solutions.}
Conversely, from an enterprise application perspective, it would be extremely useful to answer questions by means of intelligent combined uses of the input databases with other logic-intensive sources of knowledge (e.g., regulatory bodies, best practices, domain experts, etc.).
For instance, in the context of financial cases like those of interest for a Central Bank, answering questions such as \textit{``why does shareholder X exert a relevant influence on the financial intermediary Y?}'' (\textbf{explanation}), or \textit{``how does this smart contract behave?''} (\textbf{description}), or ``\textit{is the merger of banks Y and W lawful from a regulatory perspective?}'' (\textbf{question answering}), or \textit{``based on data, how many ties with other intermediaries does Z have?''} (\textbf{text-to-query translation}) would be an essential asset.

At present, all the mentioned tasks are far from being solved by off-the-shelf libraries or, directly, by most recent LLMs, and are open research.
Going into the details of each of them is beyond the scope of this paper, but the motivations laid out mainly in a question-answering perspective give the flavour of why LLMs are not enough.
It is worth remarking, though, that even the translation task, for which thanks to LLMs much progress has been made in the transformation of natural language into the target query languages (say, SQL, SPARQL, etc.)~\cite{wang2019rat,NLsparql} is still a largely unsolved problem, especially in the context of languages with an elaborate grammar and complex queries~\cite{fu2023catsql}.

\smallskip \noindent
\textbf{Ontological reasoning.}
From another perspective, in the \textit{Knowledge Representation and Reasoning}~\cite{KrotzschT16} (KRR) community, the state-of-the-art research on ontological reasoning over EKGs makes a point of being able to offer a compact combination of factual database information (the \textit{extensional knowledge}) and formally specified business awareness, for instance in the form of logical rules (the \textit{intensional knowledge}), to serve \textit{domain-specific query answering} in an accurate manner. 
For example, logical KGs exploiting efficient fragments of the \textit{Datalog$^\pm$} family~\cite{CaliGL12} have been successfully adopted for financial applications~\cite{BellomariniFGS19}. 

Yet, there is an impedance mismatch between NLP and ontological reasoning, which lacks the flexibility and the language orientation to solve explanation, description, question answering, and translation tasks: queries need to be specified in KRR formalisms; all the inputs and the results are facts/n-tuples/triples; the generation of new knowledge is possible only to the extent reasoning rules capture it.
Conversely, while being very good at manipulating human language, LLMs lack a comprehensive domain model, a pillar of KRR approaches.

\begin{figure}[t!]
    \centering
    \includegraphics[width=0.8\textwidth]{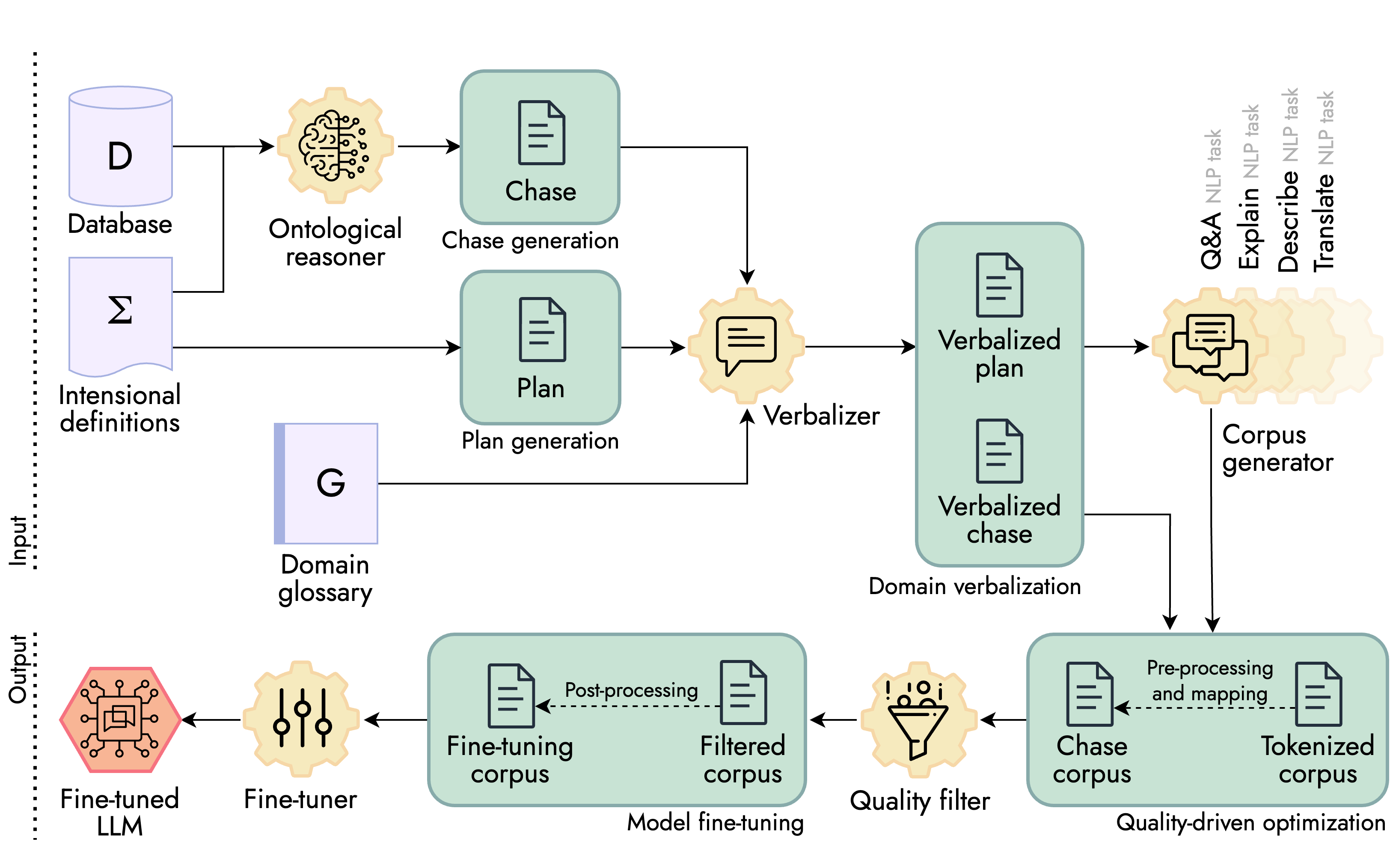}
    \caption{Neurosymbolic pipeline for reasoning-based LLM fine-tuning.}
    \label{fig:pipeline}
\vspace*{-3mm}
\end{figure}

\smallskip \noindent
\textbf{An integrated approach.}
This paper strives to strengthen LLMs in their use for task- and domain-specific applications, by letting the fine-tuning process be driven by an ontological reasoning task on an EKG.
We operate in the context of the \vadalog~\cite{BellomariniBGS22} system, a Datalog-based reasoning engine for EKGs, that finds many industrial applications~\cite{BellomariniFGS19}.
We use \vadalog to explore the factual information derived by applying the domain rules, via the \chase procedure~\cite{MaierMS79}, to the enterprise data and synthesize a fine-tuning corpus that covers the entire ``reasoning space'' to convey domain-specificity to the LLM.
A summary of the resulting \textit{fine-tuning pipeline}, provided in Figure~\ref{fig:pipeline}, will guide our discussion.

\smallskip \noindent
More in detail, our \textbf{contributions} can be summarized as follows.

\begin{itemize}[noitemsep,nolistsep,leftmargin=5mm]
    \item We present a \textbf{reasoning verbalization} technique that generates sets of prompt-response pairs from ground Datalog rules. We provide the algorithm and optimize it with a \textit{lifting technique} exploiting reasoning regularities.
    \item We deliver such an approach in a \textbf{novel neurosymbolic architecture} that fine-tunes task-specific LLMs for a set of four relevant NLP tasks, namely, \textit{explanation}, \textit{description}, \textit{question answering}, and \textit{translation}.
    \item We discuss a preliminary \textbf{proof-of-concept} confirming the validity of our approach and comparing models fine-tuned on ground and chase data.
\end{itemize}

\smallskip \noindent
\textbf{Overview.}
In Section~\ref{sec:approach} we present our architecture. A preliminary experimental validation is provided in Section~\ref{sec:poc}. We draw our conclusions in Section~\ref{sec:conclusion}.
\vspace*{-3mm}
%
%
%
%
\section{A Neurosymbolic Pipeline to Fine-tune LLMs}
\label{sec:approach}
The input blocks of the fine-tuning pipeline in Figure~\ref{fig:pipeline} are $D$ and $\Sigma$. They are, respectively, a database of domain facts and a set of reasoning rules, capturing the business dynamics. Our rules are expressed in \vadalog.
An EKG is a combination $\Sigma(D)$ of $D$ and $\Sigma$, obtained through reasoning. The set $\Sigma(D)$ is computed via the \chase~\cite{MaierMS79}: starting from $\Sigma(D)=D$, the chase augments $\Sigma(D)$ with facts derived from the application of the rules in $\Sigma$ to fixpoint.

Let us introduce our running example: a simple trading activity managed with a \textit{smart contract}~\cite{smartcontract}.
Here, $D$ contains a log over time of buy/sell orders from the traders who invest in the smart contract as well as market information, e.g., asset prices (\textit{Price}), or market shutdowns (\textit{MarketClosed}).

\begin{example} 
    \label{example:smart_contract}
    \textit{The following set $\Sigma$ contains the \vadalog rules governing the basic functioning of the market, i.e., under which conditions the orders are accepted and how profits and losses are computed.}
    {\small
    \begin{align*}
         \textit{Open}(x,y,t_1), \neg \, \textit{MarketClosed}(t_1) & \to \textit{Accepted}(x,y,t_1) \tag{$1$}\\
        \textit{Accepted}(x,y,t_1), \textit{Price}(p_1,t_1), k=y*p_1 & \to \textit{Position}(x,y,k,t_1) \tag{$2$}\\
        \textit{Close}(x,t_2), \textit{Price}(p_2,t_2), \textit{Position}(x,y,k,t_1),\\ t_2>t_1, pl=y*p_2-k & \to \textit{Return(x,pl)} \tag{$3$}
    \end{align*}
    }%
    \textit{If a trader $x$ wants to open a position (buy) on a certain asset of size $y$ at time $t_1$ and the market is open at $t_1$, the order is accepted (rule~$1$).     
    If the order by $x$ is accepted and the asset price at $t_1$ is $p_1$, then $x$ holds a position on the market at time $t_1$ of size $y$ and of notional (total value) $k$ equal to $y * p_1$ (rule~$2$).
    If, later at $t_2$, trader $x$ decides to close its  position (sell) and the price at $t_2$ is $p_2$, then $x$ gets returns (profits or losses) from its trading activity as $y*p_2 -k$ (rule~$3$).}
\end{example}

Applying the vision we laid out to Example~\ref{example:smart_contract}, the goal of our pipeline is fine-tuning an LLM to address \textit{explanation}, \textit{description}, \textit{question answering}, and

\newpage

\begin{algorithm}[t]
    \caption{Reasoning-based LLM Fine-tuning.}
    \begin{algorithmic}[1]
        \scriptsize
            \Function{ReasoningFineTuning}{$D,\Sigma,G,\mathit{model},\mathit{nlpTask}$}
                \Let{$\mathit{chase}$}{$\vadalog.\mathsf{reason}(D,\Sigma)$} \Comment{chase generation}\label{alg:line:reasoning}
                \Let{$\mathit{verbChase}$}{$\emptyset$}
                \ForEach{$\mathit{step}$ in $\mathit{chase}$} 
                    \Let{$\mathit{stepAggrContrib}$}{$\emptyset$}
                    \If{$\mathsf{hasAggregate}(\mathit{step}.\mathsf{getRule}())$}
                        \Let{$\mathit{stepAggrContrib}$}{$\mathsf{composeBack}(\mathit{step},\mathit{chase})$} \Comment{aggregates retrieval} \label{alg:line:aggregation} 
                    \EndIf
                    \Let{$\mathit{verbStep}$}{$\mathsf{verbalizeStep}(\mathit{step},\mathit{stepAggrContrib},G)$}
                    \Let{$\mathit{verbChase}$}{$\mathit{verbChase} \cup \{\mathit{verbStep}$\}} \Comment{chase verbalization} \label{alg:line:chase_verbalization}
                \EndFor
                \Let{$\mathit{verbPlan}$}{$\mathsf{verbalizePlan}(\Sigma.\mathsf{getLogicPlan}())$} \Comment{logic plan verbalization} \label{alg:line:plan_verbalization}
                \Let{$\mathit{tokenizedCorpus}$}{$\mathsf{generate}(\mathsf{preprocess}(\mathit{verbPlan},\mathit{nlpTask}))$} \Comment{tokenized corpus generation} \label{alg:line:generation}
                \Let{$\mathit{chaseCorpus}$}{$\emptyset$}
                \ForEach{$\mathit{verbStep}$ in $\mathit{verbChase}$} \Comment{chase mapping} \label{alg:line:mapping}
                    \Let{$\mathit{chasePromptResp}$}{$\mathsf{map}(\mathit{tokenizedCorpus},\mathit{verbStep})$}
                    \Let{$\mathit{chaseCorpus}$}{$\mathit{chaseCorpus} \cup \{\mathit{chasePromptResp}$\}}
                \EndFor
                \ForEach{pair $\langle \mathit{prompt},\mathit{resp} \rangle$ in $\mathit{chaseCorpus}$} \Comment{quality-driven optimization} \label{alg:line:postprocessing_start}
                    \Let{$\mathit{qualityScore}$}{$\mathsf{checkQuality}(\langle \mathit{prompt},\mathit{resp} \rangle,\mathit{nlpTask},\mathit{verbChase})$}
                    \If{$\mathit{qualityScore} \leq \mathit{threshold}$}
                        \Let{$\mathit{chaseCorpus}$}{$\mathit{chaseCorpus} \setminus \{\langle \mathit{prompt},\mathit{resp} \rangle$\}} \label{alg:line:filtering}
                    \Else
                        \Let{$\mathit{chaseCorpus}$}{$\mathit{chaseCorpus} \cup \mathsf{paraphrase}(\langle \mathit{prompt},\mathit{resp} \rangle)$} \Comment{corpus paraphrasing}
                    \EndIf
                \EndFor
                \Let{$\mathit{fineTuningCorpus}$}{$\mathsf{postprocess}(\mathit{chaseCorpus})$} \label{alg:line:postprocessing_end}
                \Let{$\mathit{ftModel}$}{$\mathsf{fineTune}(\mathit{model},\mathit{fineTuningCorpus})$} \Comment{model fine-tuning} \label{alg:line:finetuning}
                \State \Return $\mathit{ftModel}$
           \EndFunction
    \end{algorithmic}
\label{alg:pipeline}
\end{algorithm}
\vspace*{-10mm}

\noindent
\textit{text-to-query translation} tasks for the simple trading activity at hand. Let us follow Figure~\ref{fig:pipeline} and Algorithm~\ref{alg:pipeline} to describe the application of the pipeline to a database $D$ = \{\textit{Open}(\textit{EGTech},0.3,1), \textit{Open}(\textit{IEComp},0.5,1), \textit{Price}(124,1), \textit{Price}(147,9), \textit{Close}(\textit{EGTech},9), \textit{MarketClose}(5)\}.

\smallskip \noindent
\textbf{Chase generation.}
The first step of our pipeline builds the chase $\Sigma(D)$, that is, the expansion of $D$ with the facts that can be derived by applying the rules of $\Sigma$ (line~\ref{alg:line:reasoning}, in the algorithm).
Rule~$1$ generates the fact $\textit{Accepted}(\textit{EGTech},0.3,1)$, as the market is not closed at time $1$.
Then, $\textit{Position}(\textit{EGTech},0.3,37.2,1)$ is derived via rule~$2$.
Finally, as trader $\textit{EGTech}$ closes the position, i.e., sells the asset, at time $9$ and the price goes up to $147\$$, then $\textit{EGTech}$ gets a profit of $6.9\$$.

\smallskip \noindent
\textbf{Domain verbalization.} Whenever a \vadalog rule is involved in the \chase, it is translated into pure text with a deterministic transformation, based on the \textit{select-project-join} semantics, which looks up a \textit{glossary} $G$ of atom descriptions. When rules involve aggregation functions, allowed in \vadalog, the process is less straightforward and involves unfolding a chain of chase activations altogether~\cite{DBLP:journals/tcs/AfratiGT03} (line~\ref{alg:line:aggregation}).
At the end of this phase, we are in hold of a ``\textit{since-then} closure'' of our domain, that focuses on what can be obtained by activating the intensional knowledge of $\Sigma$. From another perspective, $\Sigma$ can be seen as an \textit{attention} mechanism, to select the fragment of $D$ that one wants to verbalize. 
For instance, with respect to our running example, the chase step $\mathit{Open}(\mathit{EGTech},0.3,1), \neg \mathit{MarketClose}(1) \to \mathit{Accepted}(\mathit{EGTech},0.3,1)$ (rule~1) is verbalized as: \textit{Since the trader EGTech at time 1 sends an order to open a position of size 0.3, and it is not true that 1 is a time when the market is closed, then the order of size 0.3 by EGTech is accepted at time 1}.

\begin{figure}
    \centering
    \includegraphics[width=\textwidth]{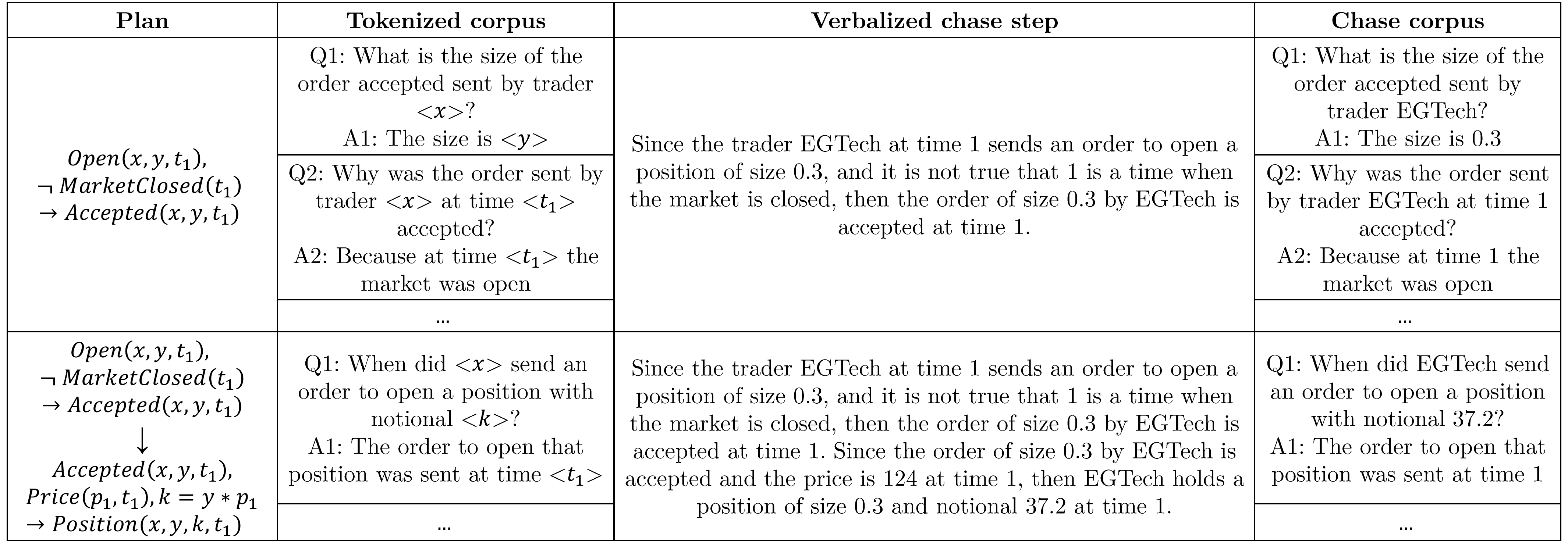}
    \caption{From plans to fine-tuning corpus, in our running example.}
    \label{fig:corpus_generation}
\vspace*{-1mm}
\end{figure}

\smallskip \noindent
\textbf{Fine-tuning corpus generation.} With the basic verbalization available, we are now ready to generate the fine-tuning corpus.
We consider the corpus generation itself as a text manipulation task and exploit the effectiveness of powerful pre-trained LLMs~\cite{brown2020language}, such as GPT-3, to synthesize a finite set of possible prompt-response pairs. Here we have two goals: 1) minimising the number of ``calls'' to the LLM, for cost- and time-efficiency reasons; 2) avoiding any ground value (coming from the EKG) being disclosed to the LLM, for data protection reasons. We leverage the regularity of logical languages and resort to a \textit{lifting technique}. We build a \textit{logic plan} out of $\Sigma$ (line 10). A plan is the equivalent in our context of a database execution plan and can be seen as the dependency graph of the rules of $\Sigma$, where nodes represent rules and edges stand for head-body dependencies. The plan is then verbalized, obtaining a text with tokens as placeholders for rule variables. Finally, a tokenized fine-tuning corpus is generated from the plan, after minor pre-processing (line~11). The form of the prompts depends on the task. 
Now, for each verbalized chase step, we look up the corresponding verbalized portion of the plan and instantiate its tokens (lines~13-15). Note that no invocations to the \textit{corpus generator} are needed in this phase. Figure~\ref{fig:corpus_generation} exemplifies the generation process in our example domain.
\vspace*{-1mm}

\noindent
\textbf{Quality-driven optimization.} The corpus undergoes a quality check where each pair is filtered according to an NLP-based scoring model in terms of specificity, plausibility, absence of bias, and other user-defined criteria. The filtered-in pairs are enhanced via \textit{NLP paraphrasing} to improve generalization and finally cleansed with additional post-processing procedures (lines~\ref{alg:line:postprocessing_start}-\ref{alg:line:postprocessing_end}).

\smallskip \noindent
\textbf{Model fine-tuning.} The refined corpus is injected into an LLM for task- and domain-specific fine-tuning (line~\ref{alg:line:finetuning}).
In the case of Q\&A, the model operates in a \textit{closed-book} approach, that is, it learns to map questions to the corresponding answers without extracting them from an input context, but rather encapsulating the knowledge of the domain into its internal parameters and weights~\cite{RobertsRS20}.
The resulting specialized model is provided to the user via API, and will act as a natural language interface to the EKG and the ontological reasoning at its foundation in a neurosymbolic fashion.
\vspace*{-3mm}
%
%
%
%
\section{Preliminary Validation via Proof-of-Concept}
\label{sec:poc}
We implemented our fine-tuning pipeline in \vadalog. 
A full-scale evaluation of our architecture is beyond the scope of this short work. Conversely, in this section, we propose a conceptual validation of the approach, by briefly showing some executions of the pipeline. We will not consider the text-to-query translation task, as its evaluation would require semantic comparison, which is beyond the scope of this work. 

For the proof-of-concept, we made use of a \textit{T5-large}~\cite{t5_large} model and considered the same domain as in Example~\ref{example:smart_contract}. To obtain a dataset that could be visually inspected to informally assess the quality of the textual answers given by an LLM fine-tuned with our pipeline, we performed a kind of 
\textit{ablation study}. 
For randomly chosen sets of sample questions, for the NLP tasks of interest, we compared the answers provided by an LLM fine-tuned only with ground facts (\textit{T5-large-ground}) and one fine-tuned with our pipeline (\textit{T5-large-chase}).
Both models were trained for $10$ epochs and with the same hyperparameters. The fine-tuning corpora and the models are made available~\cite{material}.

\begin{figure}[H]
    \centering
    \includegraphics[width=\textwidth]{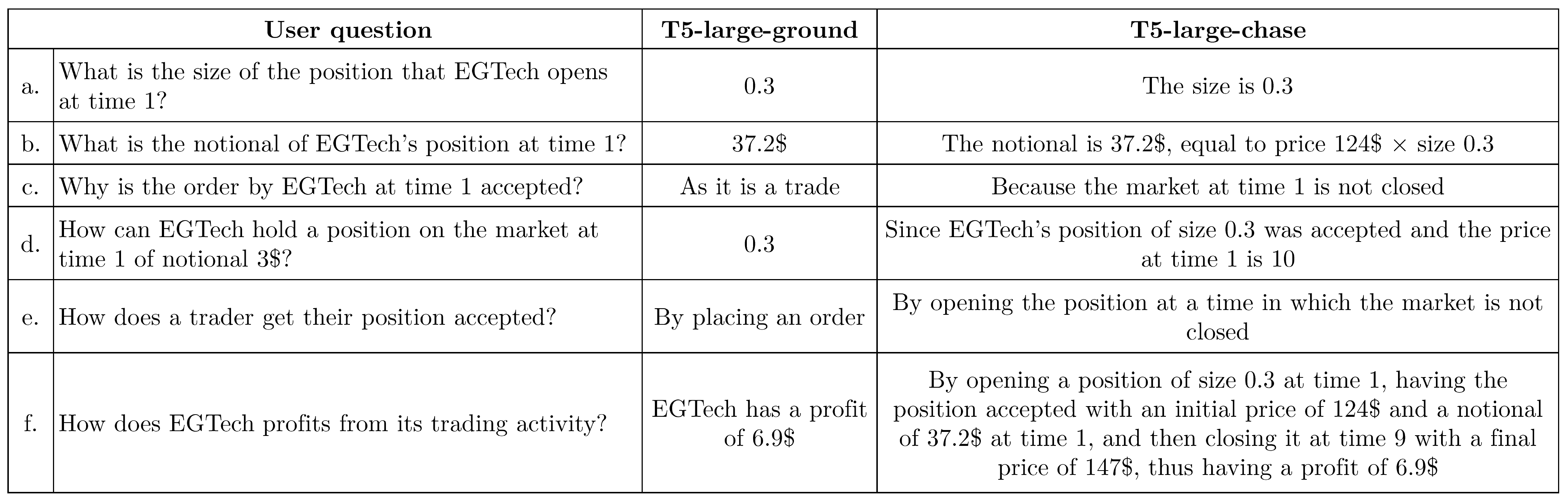}
    \caption{Proof-of-concept for our fine-tuning pipeline.}
    \label{tab:results}
    \vspace*{-4mm}
\end{figure}

Figure~\ref{tab:results} visually reports the comparison.
Questions \textit{a} and \textit{b} are the baseline, as they can be answered by facts in $D$.
Apart from a less refined write-up, the LLMs show the same output behaviour.
On the other hand, in questions \textit{c}, \textit{d}, and \textit{f} T5-large-ground is outperformed by T5-large-chase, which succeeds in answering about events related to trader \textit{EGTech}.
Actually, the corresponding facts derive from $\Sigma(D)$, which is not considered in the ground fine-tuning.
Similarly, the answer to question \textit{e} by T5-large-ground is incomplete and only T5-large-chase is able to use the specific domain knowledge from rule~$1$ of Example~\ref{example:smart_contract}.
\vspace*{-5mm}

%
%
%
%
\section{Conclusion}
\label{sec:conclusion}
\vspace*{-2mm}
According to a recent work~\cite{DBLP:conf/eacl/YuanHVKM23} appeared in the European Chapter of the Association for Computational Linguistics, pre-trained language models cannot yet perform deductive reasoning: they are still unable to generalize logical rules and, even when rules are provided, LLMs tend to forget previously inferred facts. While no extensive comparison between transformer architectures and reasoning approaches has been conducted yet, our work showed that LLM performance for domain-specific NLP tasks can be visibly improved by producing a fine-tuning corpus as a byproduct of ontological reasoning. We capitalized on our experience in deductive reasoning to offer a first step towards a neuro-symbolic platform for reasoning on enterprise knowledge graphs.

\smallskip \noindent
\textbf{Acknowledgements.}
The work on this paper was partially supported by the Vienna Science and Technology Fund (WWTF) grant VRG18-013.
%
%
%
%
%
%
\bibliographystyle{splncs04}
\bibliography{biblio}
\end{document}